\documentclass[iicol,a4paper, pdflatex,sn-nature]{sn-jnl}

\usepackage{balance}
\usepackage{microtype}   
\usepackage{dblfloatfix} 
\usepackage{graphicx}%
\usepackage{multirow}%
\usepackage{amsmath,amssymb,amsfonts}%
\usepackage{amsthm}%
\usepackage{mathrsfs}%
\usepackage[title]{appendix}%
\usepackage{xcolor}%
\usepackage{textcomp}%
\usepackage{manyfoot}%
\usepackage{booktabs}%
\usepackage{algorithm}%
\usepackage{algorithmicx}%
\usepackage{algpseudocode}%
\usepackage{listings}%
\begin{document}
\setcitestyle{super}

\title[Article Title]{Analyzing Condensation on Patterned Surfaces via

Deep Learning}


\author[1,2,3]{\fnm{Tae E.} \sur{Cooper}}

\author[2]{\fnm{Michael} \sur{Gee}}

\author[3]{\fnm{Sheila} \sur{Whitman}}

\author[4]{\fnm{Jemma} \sur{Dengel}}

\author[5]{\fnm{Anthony} \sur{Katselas}}

\author[2]{\fnm{Duygu} \sur{Dengiz}}

\author[2,6]{\fnm{Frank} \sur{Keutsch}}

\author[2,6]{\fnm{Joanna} \sur{Aizenberg}}

\affil[1]{\orgdiv{School of Electrical, Computing, and Software Engineering}, \orgname{University of Arizona}, \orgaddress{\city{Tucson}, \state{AZ}}}

\affil[2]{\orgdiv{John A. Paulson School of Engineering and Applied Sciences}, \orgname{Harvard University}, \orgaddress{\city{Cambridge}, \state{MA}}}

\affil[3]{\orgdiv{Department of Mathematics}, \orgname{University of Arizona}, \orgaddress{\city{Tucson}, \state{AZ}}}

\affil[4]{\orgdiv{Department of Biomedical Engineering}, \orgname{University of Arizona}, \orgaddress{\city{Tucson}, \state{AZ}}}

\affil[5]{\orgdiv{Department of Chemistry}, \orgname{University of Sydney}, \orgaddress{\city{Sydney}, \state{NSW}, \country{Australia}}}

\affil[6]{\orgdiv{Department of Chemistry}, \orgname{Harvard University}, \orgaddress{\city{Cambridge}, \state{MA}}}

\abstract{Dropwise condensation is key to power and water systems. State-of-the-art condensation surfaces are often inhomogeneous, either patterned by design or random defects. Inhomogeneous surface features dictate the rates of each stage of dropwise condensation–nucleation, growth, coalescence, and shedding. Characterizing these rates requires the precise detection of thousands of droplets. Yet, existing segmentation methods are unable to detect droplets on inhomogeneous surfaces.  We developed TileSAM3, a universal droplet segmentation tool built on the Segment Anything Model 3 (SAM3). Hierarchical tiling with per-tile inference followed by post-processing extended SAM3’s detection cap and raised the effective resolution of small droplets. We created two datasets: High Quality Droplet Dataset (N=365,826; HQDD) with homogeneous surfaces and Low Quality Droplet Dataset (N=63,577; LQDD) with patterned, defect-laden surfaces under adverse lighting. TileSAM3 closely reproduced thresholding annotations on HQDD, surpassed it over sixfold on LQDD (Average Precision 0.57 versus 0.088), and outperformed other deep learning models on all metrics evaluated. Combined with KD-tree tracking, TileSAM3 identified localized nucleation and coalescence in narrow active bands on a stiffness-patterned surface. TileSAM3 enables analysis of understudied inhomogeneous substrates with poor lighting and videography which facilitates faster, lower cost research of condensation materials.}


\maketitle
\section{Introduction}\label{sec1}
Dropwise condensation is a phase-change process in which vapor condenses into discrete droplets on a surface, rather than a continuous film\cite{ma2020recent}. This mode of condensation can substantially enhance heat transfer efficiency relative to filmwise condensation, and has drawn particular interest for applications in thermal power generation, chemical refining, water harvesting, desalination, refrigeration, and electronics cooling\cite{lee2012water,ghalavand2015review,mudawar2001assessment}. The efficiency of these processes is typically quantified through device-level metrics such as heat flux or collection rate\cite{elfil2020review}. These metrics, however, aggregate over the underlying kinetics of nucleation, growth, coalescence, and shedding, any of which can be rate-limiting. Classical models make heat flux an explicit function of the drop size distribution\cite{rose1973dropwise,graham1973drop}. Resolving the individual rates therefore requires precise droplet characterization, conventionally obtained by imaging the condensing surface and extracting droplet size distributions, with droplets identified through binary thresholding\cite{tancon2022simultaneous,parin2020heat,martin2017recognition,damoulakis2021quantifying} or, more recently, learned detectors\cite{chen2024deep,liu2024super,suh2021deep,lee2026analysis,ma2022condensation}. To date, demonstrations of these methods have been largely confined to liquid-infused surfaces\cite{weisensee2017condensate}, polymer brush coatings\cite{li2024durable}, and other substrates that are homogeneous at the length scale of the target droplet.

While homogeneous surfaces have been well characterized and studied, state-of-the-art condensation surfaces are patterned, whether through geometric texture\cite{miljkovic2013jumping} or chemical applications\cite{varanasi2009spatial}. Such patterns can be tuned to alter the rate of one or more condensation component: designed asymmetric bumps, for example, direct droplet growth and shedding to speed transport and refresh the surface\cite{park2016condensation}. Industrially relevant substrates, by contrast, are shaped by defects rather than design. These arise during fabrication, when randomly distributed coating defects seed nucleation\cite{enright2013condensation}, and accumulate over time as micro- and nano-structures wear and degrade\cite{xie2019dropwise}. A representative case is the shrinkage cracks that form in inverse opal coatings during drying, which favor rapid condensate transport and surface renewal\cite{adera2021enhanced}. Resolving how these features govern droplet behavior requires quantifying spatial patterns in droplet size and distribution across a surface, yet to our knowledge no current image-based method can do this on patterned or defect-laden surfaces.

Analyzing droplet populations on such inhomogeneous surfaces is inherently difficult, as surface features appear in the image alongside the droplets. Patterns and defects cause variations in contrast and blur droplet boundaries, which makes it difficult for brightness-dependent threshold-based techniques\cite{tancon2022simultaneous,parin2020heat,martin2017recognition,damoulakis2021quantifying} to reliably segment droplets from these surfaces. Specialized imaging with toroidal lighting and binary thresholding has enabled simultaneous measurement of heat flux and droplet population\cite{tancon2022simultaneous,parin2020heat}. Pillared surfaces have been addressed through a segmentation algorithm tailored to that specific pattern\cite{martin2017recognition}. Morphological operations have been applied to compensate for low image quality, yet this approach still produced droplet-count errors as high as 29\%\cite{damoulakis2021quantifying}. Motivated by the limitations of thresholding, recent research has explored the use of deep learning-based methods for droplet segmentation\cite{chen2024deep,liu2024super,suh2021deep,lee2026analysis,ma2022condensation}. Semi-supervised frameworks have been used to train convolutional neural networks to comprehensively evaluate droplet behavior during jumping-droplet condensation and frosting\cite{chen2024deep}. A super-resolution semantic segmentation model has been used to recover droplet masks from blurred, low-quality images, though for spray deposition rather than condensation\cite{liu2024super}. A custom-trained Mask R-CNN has been combined with a tracking algorithm to spatially resolve droplet statistics and heat transfer on homogeneous surfaces\cite{suh2021deep}. The Segment Anything Model enabled zero-shot droplet segmentation by tuning only inference parameters\cite{lee2026analysis}. Deep learning has also been used to detect droplets on specific patterned surfaces, such as a superhydrophobic lattice structure\cite{ma2022condensation}. To our knowledge, however, these methods have been developed for surfaces that are homogeneous or that carry a single texture, and no droplet segmentation tool has been demonstrated across arbitrary patterning, random defects, and adverse lighting.

We present TileSAM3, a universal tool for segmenting droplets across a wide range of substrate patterning, defects, and lighting conditions. The core architecture of our model is the Segment Anything Model 3 (SAM3), a vision-transformer based foundation model for instance segmentation developed by Meta AI\cite{carion2026sam3}. SAM3 is promptable with concepts (natural language or image exemplars) and was pre-trained on a large dataset of over 5 million images. Although SAM3 demonstrates strong zero-shot performance on general tasks, droplets are underrepresented in SAM3’s training set, so general-purpose recognition is unlikely to transfer directly to them. Further, adopting SAM3 for droplet segmentation is constrained by two properties of its architecture: a detection cap per inference from its detector decoder, and downsampling in its image encoder that diminishes detection of small droplets. 

\begin{figure*}[h]
\centering
\includegraphics[width=\textwidth]{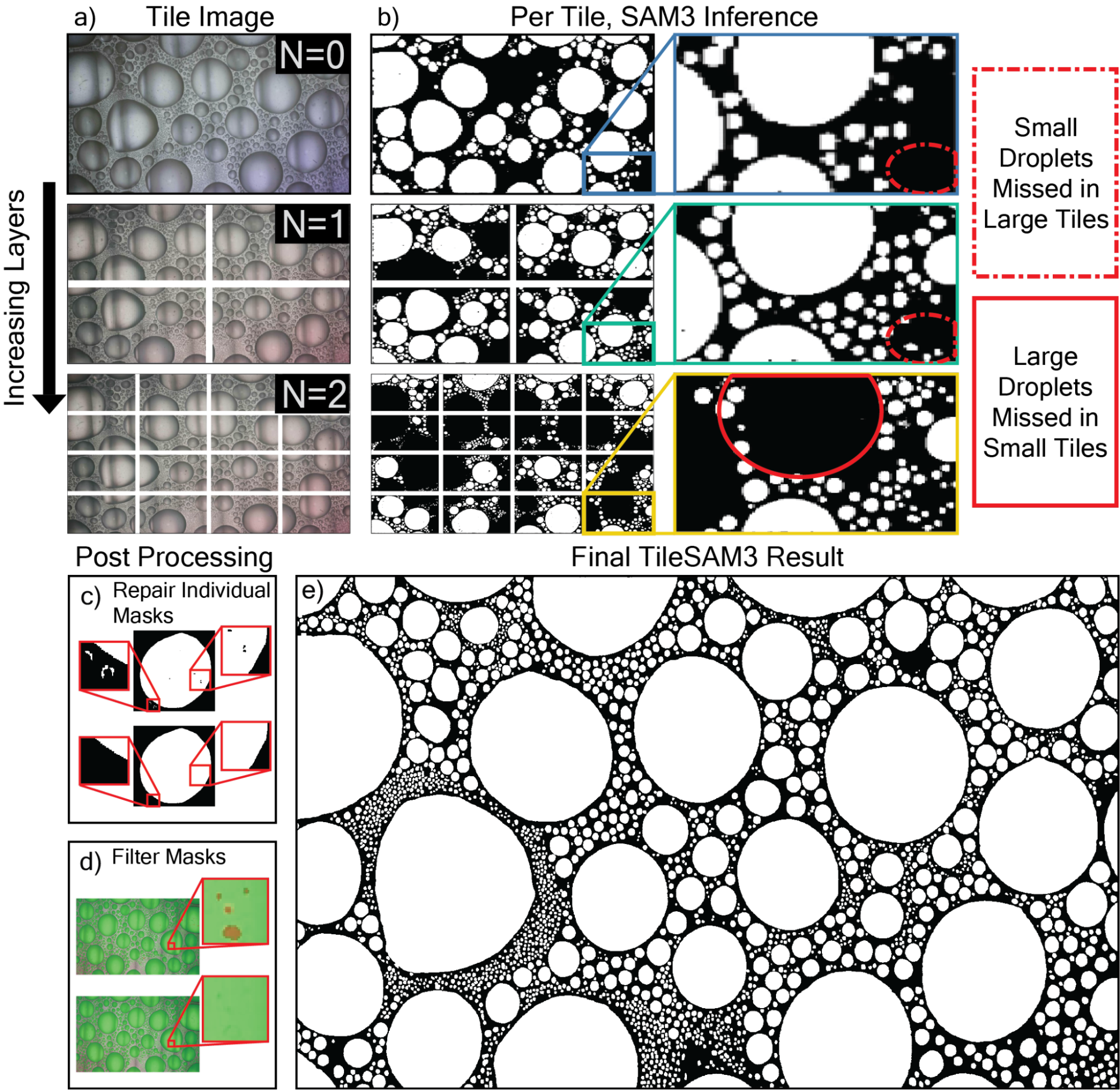}
\caption{\textbf{TileSAM3 Pipeline.} a) Images were split into N hierarchical layers of tiling. b) SAM3 inference was run once per tile (prompt = “droplet”). Droplets that were missed in one tile may have been detected in another tile over the same region in a different layer. c) Post-processing repaired masks by removing disconnected components and filling holes in masks. d) Duplicate masks between tiles and small masks fully contained within larger masks (shown in red) were removed. e) All masks were combined, which yielded a high-quality detection of droplets in the image.}\label{fig1}
\end{figure*}

To support the development of TileSAM3, two annotated datasets were created: (1) High Quality Droplet Dataset (HQDD), which enabled a comparison of models to binary thresholding on homogeneous surfaces, and (2) Low Quality Droplet Dataset (LQDD), which enabled an evaluation of segmentation performance on inhomogeneous surfaces. To overcome SAM3’s detection limit and improve small droplet detection, our TileSAM3 pipeline repeats inference on hierarchical layers of crop tiling followed by two steps of post-processing: (1) automated mask repair through hole filling and removing disconnected components and (2) non-maximum suppression to filter repetitive masks between layers. TileSAM3 outperformed all other deep learning models benchmarked on LQDD across standard instance segmentation metrics. It also generalized to HQDD after training on LQDD. Unlike binary thresholding, TileSAM3 enables the analysis of droplet dynamics on patterned surfaces. A KD-tree tracking algorithm was used to track TileSAM3 predictions between frames of droplet formation on a surface with patterned stiffness gradients. These capabilities establish TileSAM3 as a single tool that accurately segments droplets regardless of surface quality or imaging conditions, enabling spatially resolved measurements of dropwise condensation on the inhomogeneous surfaces that have been inaccessible to existing methods. 

\section{Results}\label{sec2}
\subsection{TileSAM3 Pipeline}\label{subsec1}
The core architecture of the TileSAM3 pipeline is SAM3, a foundation instance segmentation model that is prompted with natural language or image exemplars\cite{carion2026sam3}. The SAM3 detector decoder is limited to a maximum of 200 object detections per frame, which was insufficient for analysis of dropwise condensation images that often contain thousands of droplets. SAM3 is also not designed for small object segmentation, in which individual instances occupy a very low fraction of the image. Repeated inference on small regions of the image, therefore, likely improves performance by extending the detection limit and increasing the fractional occupancy of small droplets.  

In our TileSAM3 workflow, images were first split into N layers of hierarchical tiles in which each layer contains four times the number of tiles as the previous layer (Figure~\ref{fig1}a). SAM3 inference was then executed once per tile (Figure~\ref{fig1}b). A brief study on the effect of prompting schemes revealed that after fine-tuning, prompt choice had limited impact on performance, so all SAM3 segmentations were prompted with “droplet” (presented in Supporting Information i. Prompt Ablation). Smaller droplets can be detected in smaller tiles but missed in larger tiles. Larger droplets, conversely, can be detected in larger images but missed in small images. Masks were individually repaired by removing disconnected components, filling holes, and applying a convex hull to fix irregular boundaries (Figure~\ref{fig1}c). Bounding box non-maximum suppression was used to filter duplicated droplets between tiles. Masks contained within larger masks were removed to reduce false positive detections (Figure~\ref{fig1}d). Masks were then translated from tile geometry to the full image, yielding a set of droplets across size ranges (Figure~\ref{fig1}e). 

An ablation study (presented in Supporting Information ii. Tiling Depth Ablation) revealed that N=4 tiling layers yielded the highest Average Precision (AP, the mean precision-recall area over intersection over union thresholds from 0.5 to 0.95, which evaluates both detection and mask quality; see Section~\ref{subsec:metrics}). This configuration was therefore used for most experiments, with N=3 reserved for time-intensive applications (Section~\ref{subsec5}). For example, Figure~\ref{fig1}e contains 2,293 detected droplets after the full TileSAM3 pipeline (N=4), compared with only 128 detected droplets from a single SAM3 inference in the first panel of Figure~\ref{fig1}b.

\begin{figure*}[!tp]
\centering
\includegraphics[width=\textwidth,height=0.6\textheight,keepaspectratio]{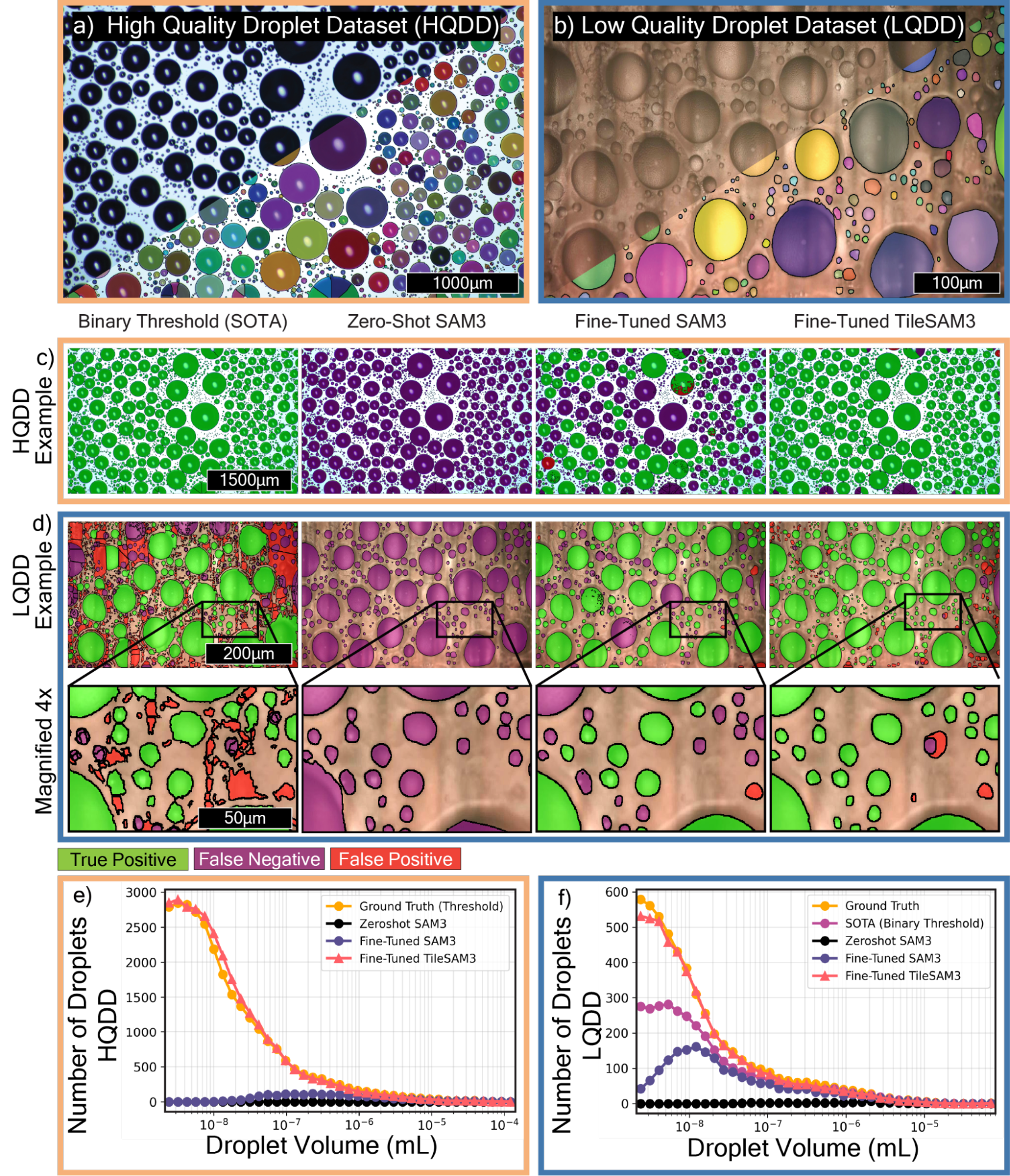}
\caption{\textbf{Overview of the datasets and comparison of fine-tuned TileSAM3 with state-of-the-art and variations.} a) High Quality Droplet Dataset (HQDD; N=365,826 annotations), an instance segmentation dataset for droplets on homogeneous surfaces under high quality imaging. On the lower right section, annotations (each identified drop is labeled by a distinct color) generated by binary thresholding are shown. b) Low Quality Droplet Dataset (LQDD; N=63,577 annotations), an instance segmentation dataset for droplets on inhomogeneous surfaces under low-quality imaging. On the lower right section, manually created ground truths by a combination of AI and human annotators are shown. c) Visual comparison of the predictions of binary thresholding (state-of-the-art; SOTA), Segment Anything Model 3 (SAM3) without fine-tuning (zero-shot), SAM3 fine-tuned on HQDD, and our TileSAM3 on an image from HQDD. SAM3 models were all prompted with “droplet.” d) Visual comparison with enlarged crops of the predictions of binary thresholding (SOTA), SAM3 without fine-tuning (zero-shot), SAM3 fine-tuned on LQDD, and our TileSAM3 on an image from LQDD.  True positive detections are shown in green; false negative detections are shown in purple; false positive detections are shown in red. e) Distributions of droplet count as a function of droplet volume from predictions on HQDD by models averaged across five-fold cross-validation. Thresholding was taken as the ground truth. f) Distributions of droplet count as a function of droplet volume from predictions on LQDD by models averaged across five-fold cross-validation. Droplet volumes were obtained by assuming droplets were spherical caps with a material dependent contact angle.}\label{fig2}
\end{figure*}

\subsection{Droplet Segmentation Datasets}\label{subsec2}
We introduced two datasets for high-quality and low-quality droplet segmentation to train and evaluate TileSAM3 across imaging conditions. 

The High Quality Droplet Dataset (HQDD) consisted of high resolution (4120 × 2644 pixels) images of droplet formation on homogeneous surfaces in which droplets have high contrast with the substrate (Figure~\ref{fig2}a). HQDD contains 365,826 instance-level droplet annotations across 96 image frames sampled from a single condensation video of a PDMS coated silicon wafer\cite{katselas2023exploring}. Ground truth annotations were generated by a computer-vision pipeline combining binary thresholding and watershed segmentation. These thresholded masks (pixel-level representations of individual droplets in images) were used as the HQDD ground truth because of the high contrast to noise ratio (CNR) of droplets (mean CNR of 4.46) produced closed masks that were representative of the true droplet shapes. HQDD was used to train and evaluate models, which provided a baseline for droplet segmentation performance under ideal conditions. 

The Low Quality Droplet Dataset (LQDD) consisted of lower-resolution (1080 × 1920 pixels) images of droplet formation on inhomogeneous surfaces, in which droplets had low CNR (mean CNR of 0.48) (Figure~\ref{fig2}b). Images in LQDD were often blurry and featured varying lighting challenges (i.e., vignette, ring, uneven). LQDD contained 63,577 instance-level droplet annotations across 96 image frames sampled from 21 videos of condensation on a variety of patterned ultrasoft bottlebrush and stiff linear PDMS (Sylgard 184) surfaces (additional information about LQDD is available in Supporting Information iii. LQDD Details). Ground-truth annotations were created through a combination of automation and manual post-processing. Candidate masks were generated by the Segment Anything Model automatic mask generator\cite{kirillov2023segment}. Manual post-processing then removed false-positive detections, refined mask shapes, and drew new masks for missed droplets larger than 10 pixels in radius. The inability of thresholding to segment droplets from non-ideal imaging necessitated this semi-manual annotation method for the creation of LQDD. LQDD is the first instance segmentation dataset for dropwise condensation on inhomogeneous surfaces under poor imaging conditions. It was used to train and evaluate models, which provided a measure of droplet segmentation performance on low-quality images. Together, HQDD and LQDD enable a broad evaluation of model performance across surface and videography.

\subsection{TileSAM3: A Universal Droplet Segmentation Tool}\label{subsec3}
To demonstrate the utility of TileSAM3 for universal droplet segmentation on both homogeneous and inhomogeneous surfaces (Figure~\ref{fig2}c--f), we compared it against three controls. Binary thresholding served as the state-of-the-art. Zero-shot SAM3, which uses the original pre-trained weights without any exposure to droplet images, tested the base foundation model's ability to recognize droplets. Fine-tuned SAM3, which was further trained on our annotated droplet images (Section~\ref{subsec:finetune}) but run as a single inference without the tiling pipeline, isolated the effect of fine-tuning from that of tiling. As demonstrated in Figure~\ref{fig2}c, TileSAM3 built upon the abilities of SAM3 and segmented droplets from high-quality images with similar performance to thresholding. Quantitative evaluation on HQDD confirmed this performance–under five-fold cross-validation, the model achieved an AP of 0.39±0.07 on all droplets and 0.94±0.02 on large droplets. As shown in Figure~\ref{fig2}d, TileSAM3 maintained its segmentation abilities on inhomogeneous surfaces and outperformed binary thresholding, which suffered from degraded mask quality due to surface imperfections. On LQDD, TileSAM3 achieved an AP of 0.57±0.02, which greatly exceeded the performance of thresholding (AP of 0.088±0.015). As shown in both Figures~\ref{fig2}c and \ref{fig2}d, zero-shot SAM3 struggled to detect droplets when prompted with “droplet,” which may be attributed to a lack of concept recognition. After fine-tuning, SAM3 recognized droplets but was hindered by its 200 object detection limit. TileSAM3 was also evaluated cross-dataset on HQDD after being trained on 80\% of LQDD (Table S2). In this experiment, TileSAM3 achieved an AP of 0.37±0.07 on all droplets and 0.93±0.01 on large droplets, falling only marginally behind the model trained on HQDD. Because HQDD and LQDD differ substantially in lighting, imaging, and surface quality, this result indicates that TileSAM3 may generalize across datasets for droplet segmentation.

As shown in Figures~\ref{fig2}e and \ref{fig2}f, TileSAM3 matched the ground-truth distribution across all droplet sizes in both HQDD and LQDD. Binary thresholding was used to annotate HQDD and was taken as the ground truth in Figure~\ref{fig2}e. Small droplets in HQDD maintained high contrast with the substrate, so thresholding could segment them reliably. The inhomogeneities and lighting conditions present in LQDD, diminished small droplets contrast with the substrate, making them especially difficult to detect for both binary thresholding and fine-tuned SAM3 (Figure~\ref{fig2}f). Both models closely matched the ground-truth curve for large droplets but substantially under-detected smaller ones. Despite TileSAM3’s distributional match, its AP for small droplets was low: only 0.29±0.07 on HQDD, rising to 0.50±0.02 on LQDD. This suggests that even when TileSAM3 recovered the correct overall size distribution, individual small-droplet masks were often of lower quality. 

\begin{figure*}[!tp]
\centering
\includegraphics[width=\textwidth]{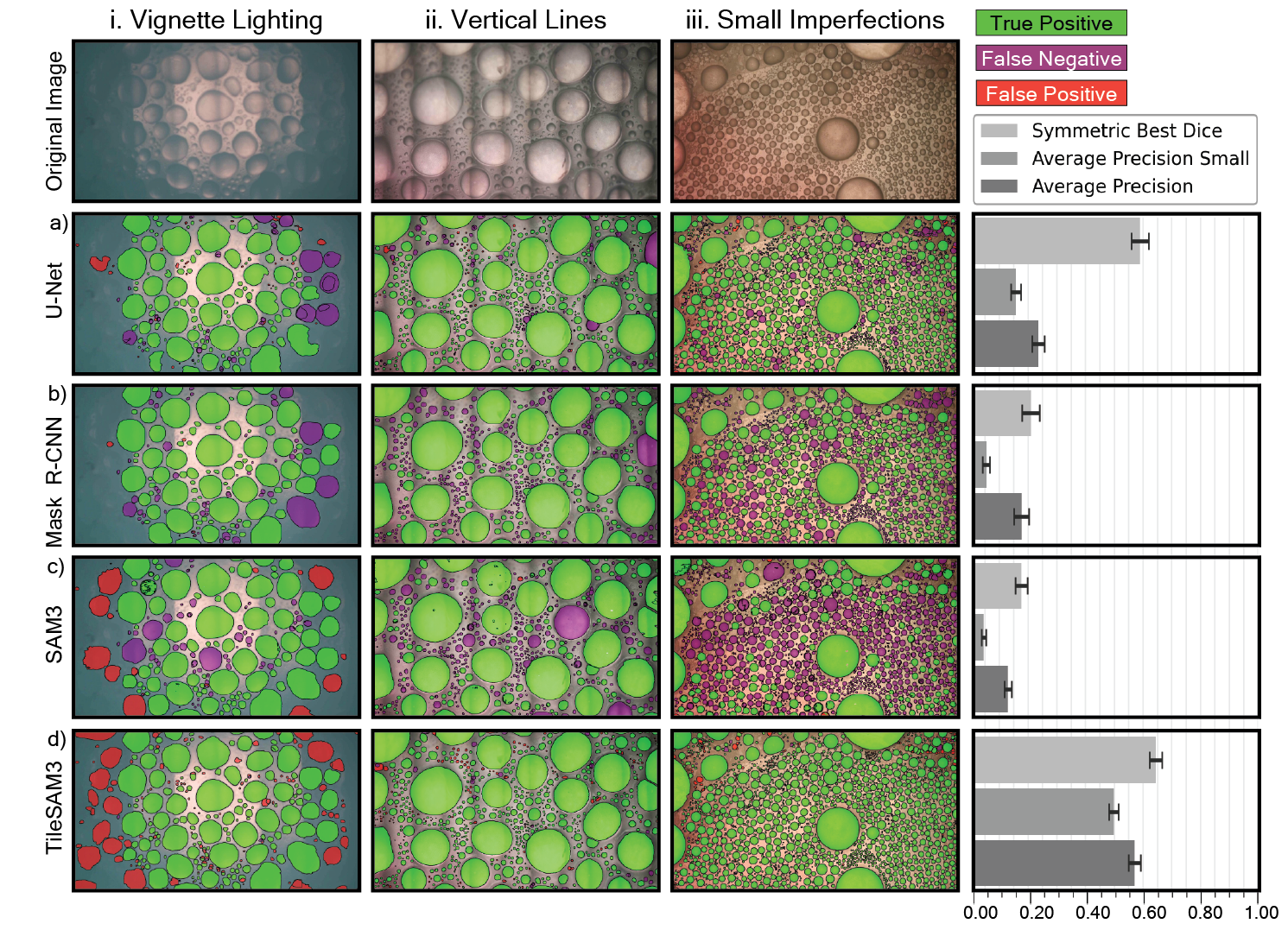}
\caption{\textbf{Qualitative and quantitative benchmark on LQDD.}  Segmentation predictions of four models: a) U-Net, b) Mask R-CNN, c) SAM3, and d) TileSAM3 on three challenging LQDD images: (i) vignette lighting that illuminates only the central region, (ii) dark vertical lines that interrupt droplet boundaries, and (iii) small surface imperfections and lighting variations that reduce local droplet contrast. The top row shows the original images. True-positive detections are shown in green, false-negative detections in purple, and false-positive detections in red. The right-hand column reports three metrics per model–Symmetric Best Dice (SBD), Average Precision for small droplets ($\mathrm{AP}_{\mathrm{small}}$), and overall Average Precision (AP)–averaged over five-fold cross-validation on LQDD.}\label{fig3}
\end{figure*}

\begin{figure*}[!tp]
\centering
\includegraphics[width=\textwidth]{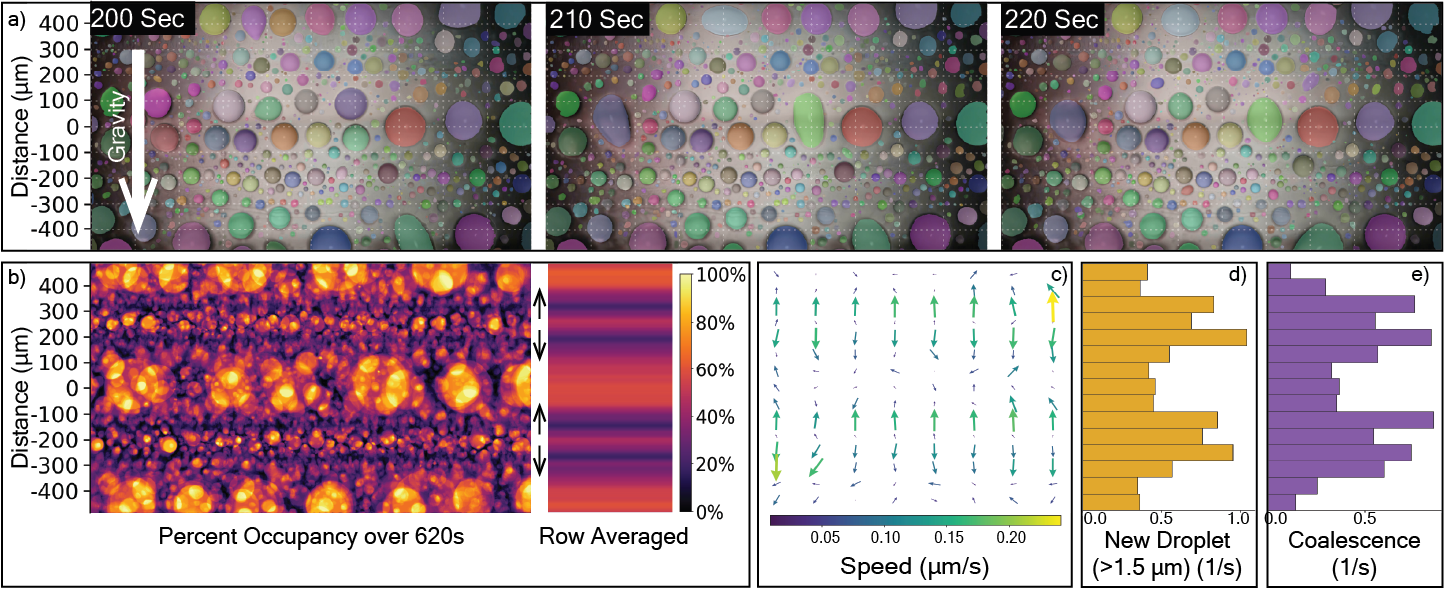}
\caption{\textbf{Droplet Tracking on a Patterned Surface.}  a) A KD-tree algorithm matched TileSAM3 predictions on frames of droplet formation in 10-second intervals. Tracking was visualized by overlaying matching droplets between frames. When droplets merged, the color of the resulting droplet was randomly chosen from the two parent droplets. b) Heatmap and row-averaged heat bar of percent occupancy over 620s of droplet formation. c) Quiver plot of local average droplet velocity across the surface. A clear pattern emerged as droplets formed in thin bands and stayed in an unstable equilibrium before migrating to larger stable regions. d) Droplet formation rate across horizontal bands of the video. The greatest amount of droplet formation occurred in the regions between stable bands. e) Coalescence rate across horizontal bands of the video. The rate of coalescence was greatest in the same regions as droplet formation–between stable bands.}\label{fig4}
\end{figure*}

\subsection{Benchmarking LQDD}\label{subsec4}
Figure~\ref{fig3} compares four segmentation models–U-Net, Mask R-CNN, SAM3, and TileSAM3–on three uniquely challenging images in LQDD, reporting both segmentation overlays and per-model dataset level performance metrics. These models were selected to provide controls for common segmentation architectures: U-Net (semantic segmentation CNN), Mask R-CNN (instance segmentation CNN), and SAM3 (vision transformer), against which TileSAM3 could be compared. In Figure~\ref{fig3}i, vignette lighting clearly illuminated only the droplets in the central region of the image. In Figure~\ref{fig3}ii, dark vertical lines spanned the substrate, interrupting droplet boundaries. In Figure~\ref{fig3}iii, the substrate was covered by small surface imperfections and lighting variations that affected local droplet contrast. U-Net is a semantic-segmentation model, so its predictions were separated into instances using connected-component labeling. Across the three LQDD samples, U-Net detected many droplets but missed some at the periphery of the illuminated region in i, as well as small droplets in both ii and iii (Figure~\ref{fig3}a). Mask R-CNN performed slightly worse than U-Net, missing more small droplets, especially in iii (Figure~\ref{fig3}b). SAM3 showed the weakest droplet detection of the models, limited to only 200 objects. Notably, the model detected the peripheral droplets in Figure~\ref{fig3}i that U-Net and Mask R-CNN missed. TileSAM3 mitigated the detection limit and segmented the most droplets correctly across all three scenarios, although it also falsely segmented several objects in Figure~\ref{fig3}i. Some of these false positives may have been extremely dimmed droplets that the annotators did not include in the ground truth. Each of the three images independently disrupted droplet contrast, rendering traditional thresholding unable to properly segment droplets from them. TileSAM's strong performance across the scenarios, therefore, provided an indication that it generalizes across inhomogeneous surfaces. The bar charts in Figure~\ref{fig3} report three complementary metrics\cite{molina2025advances}, each optimized at unity for each, averaged over five-fold cross-validation on LQDD: Symmetric Best Dice (SBD), which measures the quality of mask shape and overlap; Average Precision for small droplets ($\mathrm{AP}_{\mathrm{small}}$); and Average Precision. Full definitions are provided in Section~\ref{subsec:metrics}. A complete benchmark, additionally including binary thresholding and DeepLabv3+ and reporting AP for large droplets ($\mathrm{AP}_{\mathrm{large}}$), average best overlap, and instance F1, is provided in Table S3. TileSAM3 achieved the best score on all metrics. Its advantage was largest for $\mathrm{AP}_{\mathrm{small}}$, reaching 0.50±0.02 versus at most 0.16 for any control, an over threefold improvement. Among the controls, U-Net was the strongest. Its SBD (0.59±0.03) approached that of TileSAM3 (0.64±0.02), but this reflected mask quality of the droplets it did detect rather than a complete detection. Mask R-CNN and SAM3 trailed on every metric, with SAM3 performing the worst on the three metrics in Figure~\ref{fig3}. The full benchmark validated these results and placed binary thresholding lowest for AP (0.088±0.015) but ahead of SAM3 for all other metrics except $\mathrm{AP}_{\mathrm{large}}$. SAM3's weakness stemmed from its architecture–a fixed detection cap together with a ViT encoder that lost information about small droplets. Tiling mitigated both effects and refined the weakest model (SAM3) to create the strongest (TileSAM3).

\subsection{Droplet Dynamics on a Patterned Surface}\label{subsec5}
We demonstrated the utility of TileSAM3 by tracking and analyzing droplets on a surface with patterned stiffness gradients. A KD-tree tracking algorithm was used to track droplets from a timelapse of condensation on a patterned surface (Figure~\ref{fig4}a; see Section~\ref{subsec:tracking} for details). 

Aggregating occupancy over the full video sequence revealed strong spatial organization tied to the underlying pattern. As shown in Figure~\ref{fig4}b, the percent-occupancy heatmap and its row-averaged profile revealed that droplets did not populate the surface uniformly; instead they resolved into a set of periodic horizontal bands. Three categories of occupancy band were identified, which we defined by their position along the vertical axis of Figure~\ref{fig4}b and used consistently for all subsequent analysis: (1) low-occupancy transitional bands, spanning [-292 $\mu$m to -227 $\mu$m], [-162 $\mu$m to -97 $\mu$m], [+162 $\mu$m  to +227 $\mu$m], [+292 $\mu$m to +356 $\mu$m], with mean 27.9\% row-averaged percent occupancy; (2) high-occupancy unstable bands, spanning [-227 $\mu$m to -162 $\mu$m] and [+227 $\mu$m to +292 $\mu$m], with  mean 40.3\% row-averaged percent occupancy; and (3) high-occupancy thick stable bands, spanning [-486 $\mu$m to -292 $\mu$m], [-97 $\mu$m to +162 $\mu$m], [+356 $\mu$m to +486 $\mu$m],  with mean 49.8\% row-averaged percent occupancy. Comparing the spatially resolved occupancy with trajectory data (Figure~\ref{fig4}c) indicated a two-stage behavior: droplets first formed within the thin unstable bands, where they persisted in an unstable equilibrium, and then migrated through the transitional bands into the adjacent stable bands and accumulated. Motion in the transitional bands was strongly directed rather than diffusive: the average ratio of the mean velocity vector magnitude to the mean speed was 0.41 in transitional bands versus 0.09 in stable bands, where a value near one indicates uniform linear transport and a value near zero indicates isotropic motion. Migration out of each unstable band was approximately symmetric about the band center rather than biased downward–the net vertical velocity of droplets in the transitional bands above unstable bands was 0.16 $\mu$m/s (Standard Error = 0.005 $\mu$m/s) compared to -0.13 $\mu$m/s (Standard Error = 0.004 $\mu$m/s) of droplets in the transitional bands below unstable bands–indicating that transport was governed almost entirely by the underlying pattern rather than by gravitational drainage.

Resolving the dropwise condensation cycle by band clarified where each process was concentrated. As shown in Figure~\ref{fig4}d, droplet formation was not distributed evenly across the surface but was instead concentrated in the transitional and unstable bands. The formation rate for droplets larger than 1.5 $\mu$m was on average 0.92±0.10 1/s per bin in transitional bands and 0.72±0.05 1/s per bin in unstable bands compared with 0.44±0.08 1/s per bin in the stable bands. The transitional bands likely demonstrated the greatest formation rate as these regions were usually vacant and available for new droplets. The coalescence rate followed the same spatial trend: merging events occurred predominantly in the transitional and unstable regions (Figure~\ref{fig4}e), at 0.80±0.06 1/s per bin in transitional bands and 0.55±0.01 1/s per bin in unstable bands versus 0.34±0.16 1/s per bin in the stable bands. The droplet population in each band type differed correspondingly. The mean droplet radius was 10.0 $\mu$m in the transitional bands, 12.5 $\mu$m in the unstable bands, and 16.9 $\mu$m in the stable bands. The largest droplets on the surface resided predominantly in the stable bands (66.3\% of droplets above 25 $\mu$m radius). Together, these results indicated that the transitional zones acted as the active regions of the surface, hosting both the nucleation of new droplets and their subsequent coalescence, while the stable bands served primarily as “reservoirs” for larger droplets.

\section{Discussion}\label{sec12}
This work enables droplet-level analysis of patterned and defect-laden surfaces for the first time. Its central contribution is TileSAM3: hierarchical tiling which addresses the shortcomings of SAM3 for this task—a fixed per-inference detection cap and an encoder that over compresses small droplets. TileSAM3 converted the weakest model in the benchmark into the strongest, with an AP on LQDD (0.57±0.02), an over sixfold improvement compared with binary thresholding (0.088±0.015) and an over threefold improvement in small-droplet AP compared with the best control. 

The two datasets released here are a second contribution–to our knowledge, LQDD is the first instance-segmentation dataset for dropwise condensation on inhomogeneous surfaces. HQDD and LQDD span a range of substrate patterning, defect density, and imaging quality wide enough to support cross-dataset evaluation. The fact that a model trained only on LQDD scored within 0.02 AP of one trained on HQDD suggests TileSAM3 can be applied as a single tool across surfaces rather than retuned for each one, which in practice relaxes the imaging and surface-preparation constraints that have limited droplet-population measurements to idealized samples. The implications extend past the measurement itself. Because TileSAM3 recovers spatially resolved kinetics, it links local droplet dynamics directly to surface design. On the stiffness-patterned surface studied here, nucleation and coalescence concentrated in narrow transitional and unstable bands while mature droplets accumulated in stable bands, a partitioning of the condensation cycle that, to our knowledge, has not been demonstrated previously. The same capability may apply to surfaces with defects, where TileSAM3 offers a route to quantify how fabrication imperfections and wear-induced degradation govern droplet behavior, and, by extension, to defect detection during the mass production of condensation surfaces.

TileSAM3’s $\mathrm{AP}_{\mathrm{small}}$ was somewhat worse on HQDD (0.29±0.07) than on LQDD (0.50±0.02), despite HQDD's higher-quality imaging. This counterintuitive gap may be explained by dataset construction rather than model behavior. The 10-pixel manual-annotation threshold applied to LQDD likely excluded some of the smallest, hardest-to-score droplets from evaluation. This may have inflated LQDD's AP relative to HQDD, which included smaller droplets due to fully automated threshold annotations. This did not affect the legitimacy of the benchmarks, however, as the models were evaluated with the same data. The impact of this reduced mask quality depends on the downstream application. Small droplets contribute disproportionately to heat transfer during condensation, so imprecise small-droplet masks could meaningfully bias thermal estimates. These droplets contribute little to total condensate volume, so applications focused on collection rate are largely unaffected by this limitation. This question of application motivated a study of which TileSAM3 tiling depth best balanced detection accuracy against latency (Supporting Information ii. Tiling Depth Ablation). A second note about the datasets is that they were not fully independent of the methods evaluated. HQDD was annotated by thresholding, which was reliable because of the high contrast between droplets and substrate. But thresholding was also one of the baselines, so HQDD metrics represent agreement with the state-of-the-art rather than a measure of absolute boundary accuracy. Additionally, the LQDD labels were seeded by the SAM automatic mask generator and then manually corrected, which mitigated but did not completely remove the similarity between labels and SAM-based predictions. 

Two considerations might affect the benchmark rankings reported: (1) applying tiling to other baselines and (2) optimizing SAM3 prompting strategy. The resolution benefit of tiling (raising each droplet's fractional occupancy before encoding) is not specific to SAM3. Applying the same pipeline to CNN (U-Net, Mask R-CNN) baselines might narrow the advantage of TileSAM3, even though those models would not experience the increase in detection limit that tiling provided to SAM3's capped decoder. Part of TileSAM3's advantage is therefore specific to its backbone, and part may be a more general effect of tiling that a fuller ablation could quantify. Additionally, all SAM3 results reported were obtained using a single prompt ("droplet"). Alternative phrasings could have shifted both zero-shot and fine-tuned performance. In Supporting Information i. Prompt Ablation, we report the results of a brief study of alternative prompts and found that the prompt “bubble” maximized the AP of the zero-shot model. Fine-tuning reduced the dependence on prompting and improved performance for all prompts, however, so this step was necessary to maximize model performance.

\section{Conclusion}\label{sec13}

We introduced TileSAM3, a universal tool for droplet segmentation across substrate patterning, surface defects, and imaging conditions, together with two annotated datasets—HQDD and LQDD—that span homogeneous surfaces under ideal imaging and inhomogeneous surfaces under adverse imaging. By combining hierarchical tiling with automated mask repair and non-maximum suppression, TileSAM3 overcomes the detection limit and small-object insensitivity of its SAM3 backbone, outperforming binary thresholding and every deep learning baseline benchmarked on LQDD while matching thresholding on HQDD, and generalizing across datasets after training on a single one. Coupled with KD-tree tracking, the pipeline resolved the dropwise condensation cycle on a surface with patterned stiffness gradients, showing that nucleation and coalescence localized to narrow active bands while stable bands acted as reservoirs for mature droplets. Remaining limitations concern small-droplet mask quality, which could bias thermal estimates more than volumetric ones, and the extent to which tiling would benefit convolutional baselines, which a fuller ablation could quantify. Future work might investigate approximations for the volume of coalescing and shedding droplets, extend tracking to defect detection on mass produced surfaces, and clarify the relationship between spatially resolved droplet kinetics and heat transfer.

\backmatter

\section{Methods}\label{sec11}
\subsection{Fine-Tuning SAM3}\label{subsec:finetune}
To improve model robustness in varying imaging conditions, the following data augmentations were applied during training:
\begin{itemize}
    \item Random size cropping (ranging from 200 to 1008 pixels) and resizing via bilinear interpolation (ranging from 480 to 1008 pixels) followed by padding to 1008 × 1008 pixels.
    \item Random horizontal flipping.
    \item Color jittering with maximum adjustments of 20\% for brightness, contrast, and saturation, and 10\% for hue.
\end{itemize}
Preliminary testing indicated that while zero-shot SAM3 suffered from low droplet detection recall ($\sim$0), the few droplet masks it detected exhibited high shape quality. Therefore, the mask decoder was frozen, and training was driven by a composite loss containing bounding box (L1, gIoU) and classification (focal, presence) objectives. Training was optimized using the AdamW optimizer over 40 epochs with a batch size of 1. All training and inference were conducted on a single NVIDIA L4 GPU.

\subsection{Droplet Tracking on a Patterned Surface}\label{subsec:tracking}
The following method was used to track droplets in Section~\ref{subsec5}: time-lapsed video of dropwise condensation on an ultrasoft elastomer coated on a lenticular surface was recorded for 620 seconds, and one frame every 10 seconds was sampled. Each sampled frame was segmented using the TileSAM3 pipeline with N=3 layers of tiling. A K-dimensional tree algorithm was used to match droplets between consecutive frames based on their location, size, and eccentricity. Coalesced droplets were assigned the IDs of every previously individual droplet that contributed to the new form, and droplets were determined to have merged if they could not be identified after two frames.

\subsection{Evaluation Metrics}\label{subsec:metrics}
For a predicted mask $P$ and a ground truth mask $G$, the intersection over union (IoU) score\cite{molina2025advances} measures the degree of overlap between $P$ and $G$, and is defined as:
\begin{equation}
    \mathrm{IoU}(G, P) = \frac{\lvert G \cap P \rvert}{\lvert G \cup P \rvert}
\end{equation}

\textbf{Average Precision (AP)} is a widely used metric that measures the precision of predicted masks at various IoU thresholds. AP is the area under the precision-recall curve obtained by ranking predictions by confidence (for semantic models, all masks were ranked equally) and matching them greedily to ground truth masks at IoU threshold $t$. AP is therefore defined as:
\begin{equation}
    \mathrm{AP}(t) = \int_{0}^{1} p_t(r)\, \mathrm{d}r
\end{equation}
where $p_t(r)$ is the precision-recall curve. We report the COCO-style average:
\begin{equation}
    \mathrm{AP} = \frac{1}{10} \sum_{t \in \{0.50,\, 0.55,\, \ldots,\, 0.95\}} \mathrm{AP}(t)
\end{equation}
We also report $\mathrm{AP}_{\mathrm{small}}$ and $\mathrm{AP}_{\mathrm{large}}$, the AP of masks with fewer than $32^2$ pixels and more than $96^2$ pixels, respectively.

\textbf{Symmetric Best Dice (SBD)} is a confidence-agnostic metric that reports the accuracy of predicted mask shape and overlap. SBD is the minimum of the best dice score between predicted and ground truth masks and vice versa. The best dice score is defined as:
\begin{equation}
    \mathrm{BD}(y^a, y^b) = \frac{1}{M} \sum_{i=1}^{M} \max_{1 \le j \le N} \frac{2\,\lvert y_i^a \cap y_j^b \rvert}{\lvert y_i^a \rvert + \lvert y_j^b \rvert}
\end{equation}
where $y^a$ and $y^b$ are two sets of masks with $M$ and $N$ elements, respectively, and the operator $\lvert \cdot \rvert$ returns the number of pixels in a mask. SBD is therefore defined as:
\begin{equation}
    \mathrm{SBD}(y^p, y^g) = \min\left(\mathrm{BD}(y^g, y^p),\, \mathrm{BD}(y^p, y^g)\right)
\end{equation}
where $y^p$ are the predicted masks and $y^g$ are the ground truth masks.

\textbf{Average Best Overlap (ABO)} is the mean quality of the best prediction for each ground truth mask. Unlike SBD, ABO is insensitive to false positives and indicates the optimal model performance. ABO is defined as:
\begin{equation}
    \mathrm{ABO}(y^p, y^g) = \frac{1}{M} \sum_{i=1}^{M} \max_{1 \le j \le N} \mathrm{IoU}(y_i^g, y_j^p)
\end{equation}

\textbf{Instance F1 (InsF1)} provides a balanced evaluation of the precision and recall of detections, giving a measure of the completeness of predictions. InsF1 is defined as:
\begin{equation}
    \mathrm{InsF1} = 2 \cdot \frac{\mathrm{InsPr} \cdot \mathrm{InsRe}}{\mathrm{InsPr} + \mathrm{InsRe}}
\end{equation}
where $\mathrm{InsPr}$ and $\mathrm{InsRe}$ are the instance precision and recall, respectively.

\section*{Supplementary Information}
Supplementary information will be made available upon publication, including the prompt ablation (Figures S1--S3), tiling depth ablation (Table S1), LQDD details, dataset ablation (Table S2), and full LQDD benchmark (Table S3).

\section*{Acknowledgments}

SEW acknowledges support from the National Science Foundation Graduate Research Fellowship Program under Grant Numbers DGE-2137419, AGS-2150290, DUE2150530, and DUE2150532. The views and conclusions contained herein are those of the authors and should not be interpreted as necessarily representing the official policies or endorsements, either expressed or implied, of the National Science Foundation.

\section*{Declarations}
\bmhead{Data and Code Availability}
Datasets and code will be made available upon publication.

\bibliography{sn-bibliography}

\end{document}